\documentclass[conference]{IEEEtran}
\IEEEoverridecommandlockouts

\usepackage{cite}
\usepackage{amsmath,amssymb,amsfonts}
\usepackage{algorithmic}
\usepackage{graphicx}
\usepackage{textcomp}
\usepackage{xcolor}
\usepackage{graphics} %
\usepackage{epsfig} %
\usepackage{mathptmx} %
\usepackage{times} %
\usepackage[misc,geometry]{ifsym}
\usepackage{multirow}
\usepackage{hyperref}
\def\BibTeX{{\rm B\kern-.05em{\sc i\kern-.025em b}\kern-.08em
    T\kern-.1667em\lower.7ex\hbox{E}\kern-.125emX}}
\begin{document}

\title{MedConv: Convolutions Beat Transformers on Long-Tailed Bone Density Prediction}

\author{Xuyin Qi$^{1,2,*,\dag}$,
Zeyu Zhang$^{1,3,*,\dag,\ddag}$, %
Huazhan Zheng$^{4,*}$, %
Mingxi Chen$^{5}$,  %
Numan Kutaiba$^6$, %
Ruth Lim$^{6,7}$, %
Cherie Chiang$^7$,\\ %
Zi En Tham$^8$, %
Xuan Ren$^{2}$, %
Wenxin Zhang$^{9}$, %
Lei Zhang$^{9}$, %
Hao Zhang$^{9}$, %
Wenbing Lv$^{10}$, %
Guangzhen Yao$^{11}$, %
Renda Han$^{12}$,\\ %
Kangsheng Wang$^{13}$, %
Mingyuan Li$^{14}$, %
Hongtao Mao$^{15}$, %
Yu Li$^{16}$, %
Zhibin Liao$^2$, %
Yang Zhao$^{8,\text{\Letter}}$, %
Minh-Son To$^1$ %
\thanks{$^{*}$Equal Contribution. $^{\text{\Letter}}$Corresponding author: \href{mailto:y.zhao2@latrobe.edu.au}{y.zhao2@latrobe.edu.au}}
\thanks{$^\dag$Work done while Zeyu Zhang is a researcher assistant at Flinders University.}
\thanks{$^\ddag$Project lead.}%
\thanks{$^1$Flinders University $^2$The University of Adelaide $^3$The Australian National University $^4$Zhejiang University of Technology $^5$Guangdong Technion – Israel Institute of Technology $^6$Austin Health $^{7}$The University of Melbourne $^{8}$La Trobe University $^9$University of Chinese Academy of Sciences $^{10}$Yunnan University $^{11}$Northeast Normal University $^{12}$Hainan University $^{13}$Univeristy of Science and Technology Beijing $^{14}$Hebei University of Technology $^{15}$Central China Normal University $^{16}$Hubei University}}

\maketitle

\begin{abstract}
Bone density prediction via CT scans to estimate T-scores is crucial, providing a more precise assessment of bone health compared to traditional methods like X-ray bone density tests, which lack spatial resolution and the ability to detect localized changes. However, CT-based prediction faces two major challenges: the high computational complexity of transformer-based architectures, which limits their deployment in portable and clinical settings, and the imbalanced, long-tailed distribution of real-world hospital data that skews predictions. To address these issues, we introduce \textbf{MedConv}, a convolutional model for bone density prediction that outperforms transformer models with lower computational demands. We also adapt Bal-CE loss and post-hoc logit adjustment to improve class balance. Extensive experiments on our AustinSpine dataset shows that our approach achieves up to \textbf{21\%} improvement in accuracy and \textbf{20\%} in ROC AUC over previous state-of-the-art methods.
Code will be available at \url{https://github.com/Richardqiyi/MedConv}.
\end{abstract}


\section{Introduction}
Bone health, crucial for mobility, fracture prevention, and overall well-being, is particularly important for aging populations or those with osteoporosis, a common skeletal disease that compromises bone strength by causing low bone mass and microarchitectural deterioration. This condition, increasing the risk of fragility fractures from low-energy impacts, often affects critical areas like the spine, hip, and wrist, significantly reducing quality of life ~\cite{lupsa2015bone}. Predicting bone density through CT scans to estimate T-scores offers a more precise and detailed assessment of bone health compared to traditional methods like X-ray bone density tests, which have lower spatial resolution and limited ability to detect localized bone changes. CT-based assessments can measure volumetric bone mineral density (BMD) and provide three-dimensional imaging, allowing for a comprehensive evaluation of bone quality. Studies have demonstrated that deep learning models applied to CT images can accurately predict BMD and T-scores, enhancing the detection and management of osteoporosis ~\cite{sato2022deep}. Additionally, quantitative computed tomography (QCT) has been shown to be a superior method for diagnosing osteoporosis and predicting fractures when compared to dual-energy X-ray absorptiometry (DXA) ~\cite{mori2024advancing}. These advancements highlight the potential of CT imaging in providing detailed insights into bone health, surpassing the capabilities of traditional X-ray-based methods. Recent advances in representation learning ~\cite{ji2024sine} and dense prediction \cite{zhang2023thinthick,wu2023bhsd,tan2024segstitch,ge2024esa,zhang2024meddet,cai2024msdet,tan2024segkan,zhang2025gamed}, particularly in the domain of medical imaging ~\cite{zhang2024jointvit, wu2024xlip,hiwase2025can,zhao2024landmark,cai2024medical,qi2025projectedex}, have significantly enhanced the accuracy and automation of osteoporosis detection. These advancements facilitate early diagnosis and timely intervention, providing a foundation for more effective personalized treatment and prevention strategies.

\begin{figure}[t]
    \centering
    \includegraphics[width=\columnwidth]{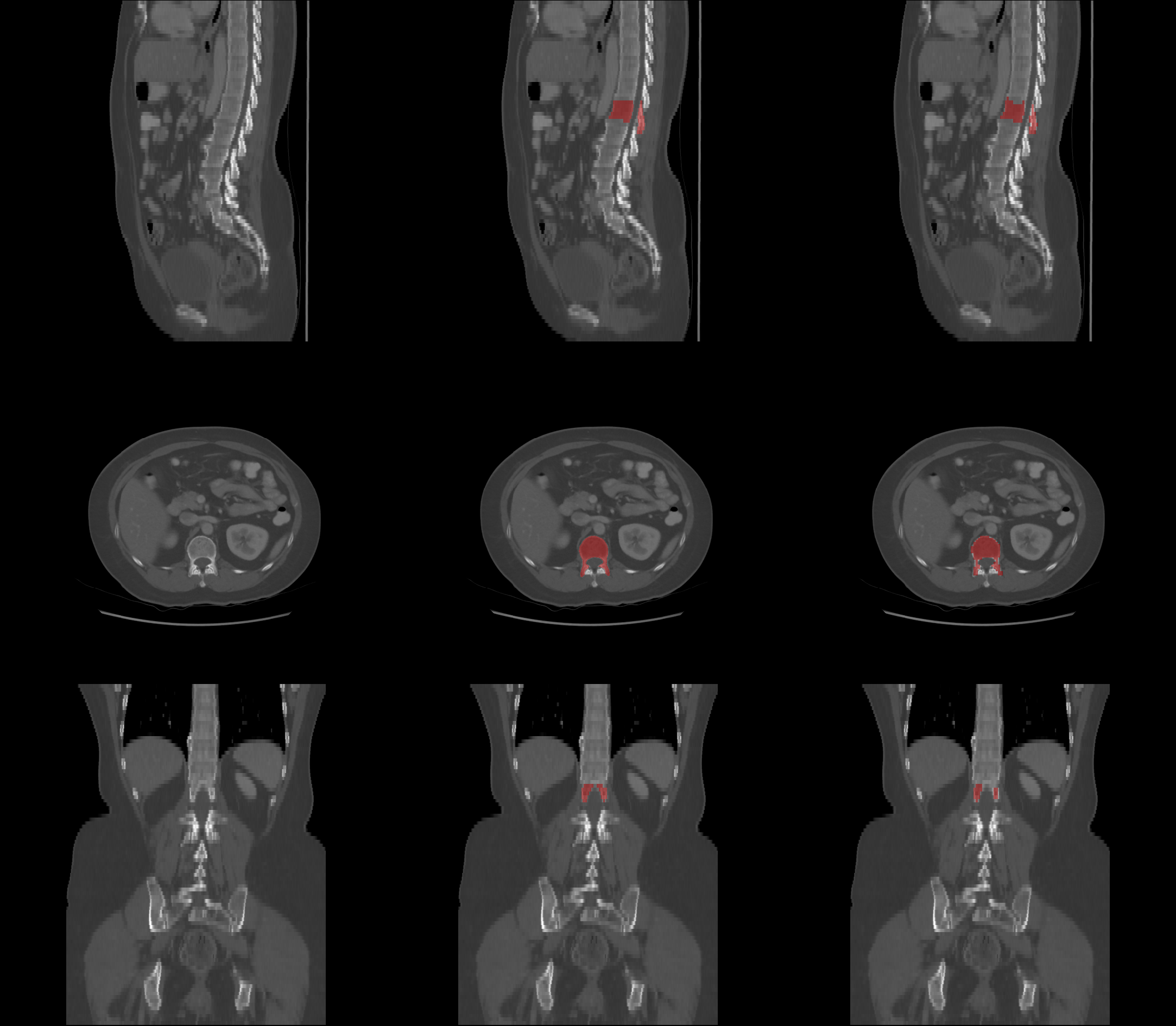}
    \caption{Visualization of segmentation results on CT images. 
    The first column shows the original images. 
    The second column represents the segmentation results from CTSpine1K~\cite{deng2021ctspine1k}. 
    The third column displays the segmentation results from TotalSegmentator~\cite{wasserthal2023totalsegmentator}. 
    Rows correspond to different anatomical planes: the sagittal plane (S) in the first row, the axial plane (A) in the second row, and the coronal plane (C) in the third row. 
    The region highlighted in red corresponds to the L5 vertebra, which plays a crucial role in diagnosing conditions like osteoporosis.}
    \label{fig:segmentation}
\end{figure}

However, predicting bone density from CT scans poses two significant challenges that hinder the effective application of advanced deep learning models.
First, transformer-based architectures, which have gained popularity in recent years for their superior performance in various domains, tend to suffer from quadratic complexity in their self-attention mechanisms. This results in substantial computational demands, particularly for high-resolution medical images like CT scans, where the input size can be extremely large. Such resource-intensive requirements make these models inefficient for deployment on portable or edge devices, which are increasingly sought after in modern healthcare for their potential to enable point-of-care diagnostics. Furthermore, the computational burden limits their feasibility in real-world clinical practice, where rapid processing and cost-effectiveness are critical.
Second, real-world hospital data often exhibits an imbalanced, long-tailed distribution, heavily skewed toward more prevalent cases of osteoporosis while containing significantly fewer samples of less common conditions, such as borderline or early-stage bone density anomalies. This data imbalance poses a considerable challenge for model training, as standard machine learning algorithms tend to prioritize the majority class, leading to suboptimal performance in predicting rare cases. Addressing this issue requires sophisticated techniques, such as class rebalancing strategies, data augmentation, or the use of domain-specific loss functions, to ensure that models can achieve robust and fair predictions across the full spectrum of cases ~\cite{johnson2019survey}.

To adress these problems, our paper presents three main contributions:

\begin{itemize}
    \item We introduce \textbf{MedConv}, a robust model that revisits convolutional approaches for bone density prediction on spinal CT scans, achieving superior performance over transformer-based models with reduced computational complexity.
    \item To address the long-tailed prediction challenge, we customize Bal-CE loss and post-hoc logit adjustment for improved class balance and accuracy.
    \item We evaluated our methods through extensive experiments on our AustinSpine dataset, applying various preprocessing techniques, which yielded improvements of up to 21\% in accuracy and 20\% in ROC AUC compared with previous state-of-the-art methods.
\end{itemize}

\section{Related Work}

\subsection{Deep Learning for Bone Mineral Density Prediction}

The prediction of bone mineral density (BMD) and the evaluation of fracture risk through the application of deep learning techniques \cite{zhang2024deep} have garnered increasing attention in recent years. A notable study by Hsieh et al. (2021) \cite{hsieh2021automated} introduced an innovative approach that utilizes deep learning models applied to plain radiographs for the automated prediction of BMD and fracture risk assessment. Their method demonstrated highly promising results, achieving area under precision-recall curve (AUPRC) scores of 0.89 for hip osteoporosis and 0.83 for spine osteoporosis prediction. Furthermore, their model exhibited an impressive accuracy of 91.7\% in estimating the risk of hip fractures. Leveraging a large dataset comprising pelvis and lumbar spine radiographs, the study underscored the potential of deep learning in addressing osteoporosis detection, particularly in scenarios where dual-energy X-ray absorptiometry (DXA) remains underutilized.

In another significant contribution, Yasaka et al. (2020) \cite{yasaka2020prediction} explored the use of CT imaging for BMD prediction, employing a convolutional neural network (CNN) specifically designed to estimate lumbar vertebrae BMD from unenhanced CT scans. Their findings demonstrated a strong correlation between CNN-predicted BMD values and those obtained through DXA, achieving area under the receiver operating characteristic curve (AUC) scores of 0.965 and 0.970 for internal and external validation datasets, respectively. This study laid the groundwork for using CT imaging as an effective alternative to DXA in BMD prediction, illustrating the capability of CNN-based models to accurately capture bone density-related features.

\begin{figure}[t]
    \centering
    \includegraphics[width=0.8\linewidth]{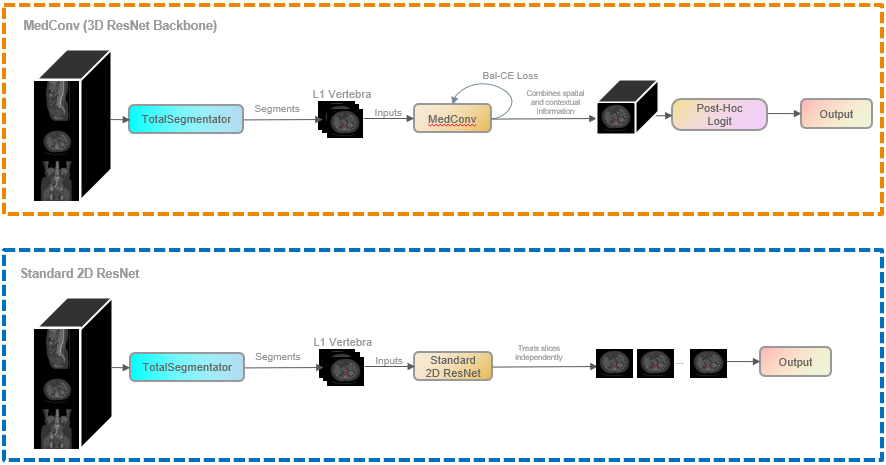}
    \caption{
        Comparison between 3D ResNet and 2D ResNet architectures for volumetric medical data processing. 
        The upper pipeline illustrates the 3D ResNet-based MedConv model, which leverages three-dimensional convolutions to capture spatial and contextual information across volumetric CT scans. 
        The inclusion of Bal-CE Loss further refines the model's focus on imbalanced data distributions, ensuring accurate predictions for the L1 vertebra segmentation task.
        Conversely, the lower pipeline showcases the standard 2D ResNet approach, where slices are treated independently without spatial continuity across adjacent slices, potentially limiting performance in tasks requiring volumetric context. 
        This figure highlights the architectural and methodological differences, emphasizing the advantages of 3D ResNet for tasks that demand structural and contextual understanding of medical images.}
    \label{fig:comparison_models}
\end{figure}

Building upon this research, Dagan et al. (2019) \cite{dagan2020automated} developed a model aimed at predicting fracture risk based on routine CT scans, particularly when DXA-derived data is unavailable. Their CT-based method demonstrated superior AUC scores and sensitivity compared to the FRAX tool when BMD inputs were excluded, indicating that CT scans can serve as a reliable resource for assessing fracture risk. This approach suggests that CT imaging could effectively compensate for the underutilization of DXA in clinical settings.

In another noteworthy study, González et al. (2018) \cite{gonzalez2018deep} proposed a direct image-to-biomarker prediction approach. By employing a deep learning regression model, they predicted BMD directly from CT scans. Their results highlighted the effectiveness of a single convolutional neural network in simultaneously segmenting relevant anatomical regions and predicting BMD values with high accuracy. This streamlined approach provides an efficient alternative to traditional methods that rely on separate segmentation and prediction steps.

Lastly, Fang et al. (2020) \cite{fang2021opportunistic} demonstrated the potential of multi-detector CT imaging for opportunistic osteoporosis screening. By combining U-Net for vertebral segmentation with DenseNet-121 for BMD estimation, their method achieved a strong correlation with quantitative computed tomography (QCT) benchmarks. This fully automated pipeline showcased the feasibility of integrating CT-derived BMD analysis into routine clinical practice for opportunistic screening. Their study highlighted how deep learning can facilitate cost-effective and automated osteoporosis detection in diverse healthcare environments.

\subsection{Addressing Long-Tailed Distribution in Classification Tasks}

Long-tailed distributions, characterized by a few dominant classes and a large number of underrepresented classes, pose significant challenges in classification tasks. These challenges arise due to the imbalance in the data distribution, which can lead to biased model predictions favoring majority classes while neglecting minority ones. Two widely adopted strategies for addressing this issue are resampling methods and balanced augmentation (BalAug), both of which aim to mitigate the effects of data imbalance by adjusting the training process.

Resampling methods involve manipulating the class distribution in the training dataset. Oversampling techniques, such as random duplication or Synthetic Minority Over-sampling Technique (SMOTE), increase the representation of minority classes, thereby providing the model with more exposure to these underrepresented categories. However, these approaches may lead to overfitting on the minority classes due to repeated exposure to the same data points. On the other hand, undersampling methods reduce the number of majority class samples to balance the dataset, but this can result in a loss of valuable information from the majority classes, as noted in ~\cite{bellinger2020remix}. Consequently, while resampling methods are straightforward and often effective, they require careful tuning to avoid introducing new challenges.

Balanced augmentation (BalAug) offers an alternative approach by integrating data augmentation techniques with class balancing. Augmentation strategies such as rotation, cropping, flipping, and other transformations are selectively applied to the minority classes, enhancing the diversity of training data for these underrepresented categories. For instance, ~\cite{cui2019class} introduced a class-balanced loss that dynamically weights samples based on their effective number, ensuring that the model learns equitably from all classes. Furthermore, advanced techniques like class-aware sampling combined with augmentation, as proposed in ~\cite{liu2022long}, have demonstrated improved performance on long-tailed datasets by carefully balancing the sampling probabilities and incorporating diverse transformations. These methods not only enrich the training data but also help the model generalize better to unseen data.

In addition to data-focused strategies, training optimization methods have emerged as powerful tools for addressing long-tailed distributions. Foret et al. (2020) ~\cite{foret2020sharpness} introduced Sharpness-Aware Minimization (SAM), a novel optimization approach designed to enhance model generalization by simultaneously minimizing the loss value and the sharpness of the loss landscape. SAM identifies parameter regions with consistently low loss, effectively mitigating overfitting and improving generalization, particularly in overparameterized models. Through rigorous evaluation on benchmark datasets like CIFAR ~\cite{krizhevsky2009learning} and ImageNet ~\cite{deng2009imagenet}, SAM demonstrated superior performance, excelling in robustness to label noise and training stability, making it a valuable addition to the arsenal of techniques for long-tailed datasets.

Building on these ideas, Fang et al. (2023) ~\cite{defazio2024road} proposed a schedule-free optimization framework to address long-tailed distributions by replacing traditional learning rate schedules with momentum-driven primal averaging. Their approach dynamically balances gradient updates, avoiding the gradient collapse often observed in imbalanced datasets. This innovative method achieved state-of-the-art results across various tasks, including CIFAR-10 and ImageNet, by combining robust convergence properties with efficient generalization capabilities. By reducing reliance on extensive hyperparameter tuning, this approach offers a practical solution for training on long-tailed data.

In summary, addressing the challenges posed by long-tailed distributions typically requires a combination of data-level and training-level strategies. Data-level approaches, such as resampling and balanced augmentation, aim to correct the imbalance in the dataset, ensuring that all classes are adequately represented during training. Training-level techniques, like SAM and schedule-free optimization, focus on improving model generalization by optimizing the training process itself. When these methods are combined effectively, they can complement each other, leveraging the strengths of both data and training interventions to achieve robust and unbiased performance on long-tailed datasets.

\section{Methodology}

\subsection{Overview}

\begin{figure}[h]
    \centering
    \includegraphics[width=\columnwidth]{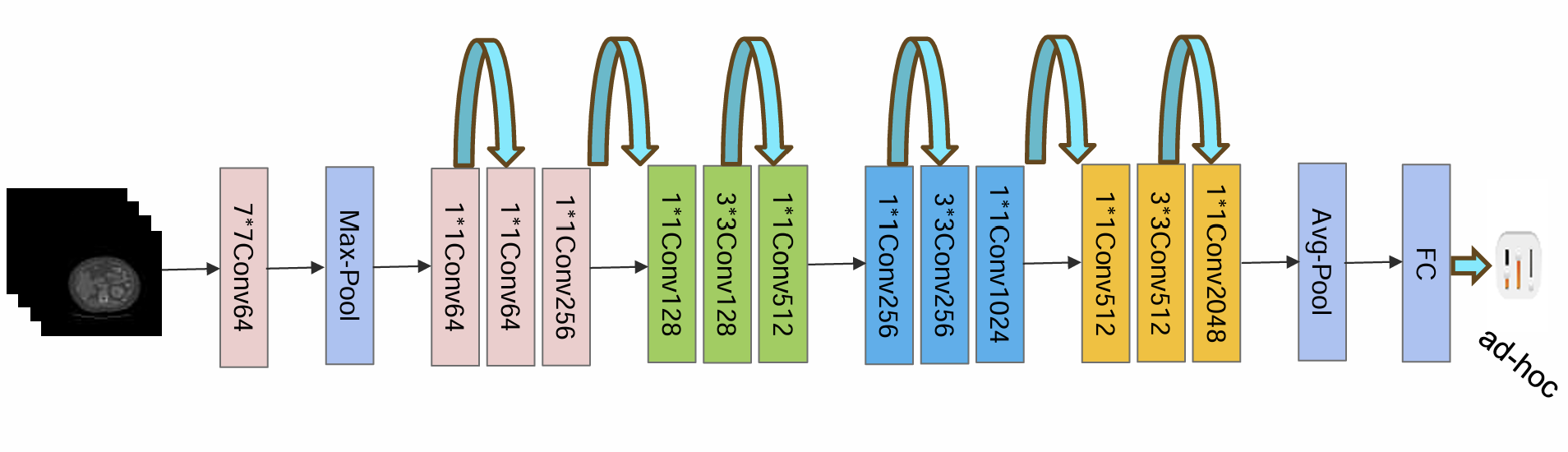}
    \caption{Architecture of the proposed MedConv model, based on a 3D ResNet-50 backbone. 
    The model leverages the volumetric spatial representation capabilities of 3D convolutions, essential for accurate bone density estimation. 
    Key methodologies include the use of Balanced Cross-Entropy (Bal-CE) loss and post-hoc logit adjustment with hyperparameters $\tau_1 = 1$ and $\tau_2 = 0.5$, which enhance class balance and calibration.}
    \label{fig:main-graph}
\end{figure}

Our proposed model is built upon a 3D ResNet-50 backbone, selected for its superior ability to capture the spatial and contextual information embedded in volumetric medical data. Unlike conventional 2D convolutional neural networks (CNNs) that process individual image slices independently, thereby neglecting depth information, the 3D ResNet-50 employs three-dimensional convolutional operations. This design enables the model to effectively encode spatial continuity within volumetric datasets such as CT scans, a critical aspect for accurate bone density prediction.

The architecture leverages residual connections to address the vanishing gradient problem, facilitating the training of deep networks while maintaining representational efficiency. Additionally, the bottleneck structure within the 3D ResNet-50 reduces computational overhead without compromising its capacity to model complex patterns inherent in high-resolution medical images.

While transformer-based architectures excel in capturing long-range dependencies and global contextual features, their computational complexity grows quadratically with input size. This limitation poses significant challenges for processing high-resolution volumetric data in resource-constrained settings. By contrast, the 3D ResNet-50 achieves an effective trade-off between computational efficiency and representational power, making it a practical and scalable choice for clinical applications.

This backbone forms the foundation of our model, providing a framework that balances accuracy and efficiency for the analysis of volumetric medical data. Its ability to integrate three-dimensional spatial information ensures robust performance, particularly in tasks requiring detailed structural understanding, such as bone density prediction.

\subsection{Balanced Cross-Entropy (Bal-CE) Loss}

To address the inherent challenges of class imbalance in bone density prediction, we adopt a Balanced Cross-Entropy (Bal-CE) loss function. Medical imaging datasets often exhibit a long-tailed distribution, with underrepresented classes being critical for diagnosis. The Bal-CE loss function is designed to emphasize these minority classes by assigning class-specific weights, \( w_i \), during training. Its formulation remains as follows:

\[
\mathcal{L}_{\text{Bal-CE}} = - \frac{1}{N} \sum_{i=1}^{N} w_i \left( y_i \log(\hat{y}_i) + (1 - y_i) \log(1 - \hat{y}_i) \right),
\]

where \( y_i \) and \( \hat{y}_i \) represent the ground truth labels and predicted probabilities, respectively. The weight \( w_i \) is dynamically computed based on the inverse frequency of each class, ensuring greater emphasis on minority classes. This targeted adjustment helps the model avoid bias toward majority classes, leading to more balanced and reliable predictions.

\subsection{Post-Hoc Logit Adjustment}

To further enhance model calibration and refine class probabilities, we introduce a post-hoc logit adjustment technique. This method applies temperature scaling to logits, fine-tuning the relative contributions of majority and minority classes. The adjusted probabilities are calculated as:

\[
\hat{y}_i = \frac{e^{z_i / \tau_1}}{e^{z_i / \tau_1} + e^{z_j / \tau_2}},
\]

where \( z_i \) and \( z_j \) denote the logits for classes \( i \) and \( j \), respectively. The temperature parameters \( \tau_1 = 1 \) and \( \tau_2 = 0.5 \) are empirically chosen to achieve an effective balance. The lower value of \( \tau_2 \) amplifies the influence of minority class logits, while \( \tau_1 \) maintains the contribution of majority classes. This mechanism mitigates the impact of class imbalance by reshaping the probability distribution, allowing the model to produce well-calibrated predictions.

The combination of the Bal-CE loss and logit adjustment strategies ensures that our model effectively learns from imbalanced datasets while maintaining robustness in clinical scenarios. Together, these methods address the challenges posed by uneven class distributions and improve the reliability of the system for bone density prediction tasks.

\section{Dataset and Evaluation Matrices}
\subsection{AustinSpine Dataset}

\begin{figure}[h]
    \centering
    \includegraphics[width=0.5\linewidth]{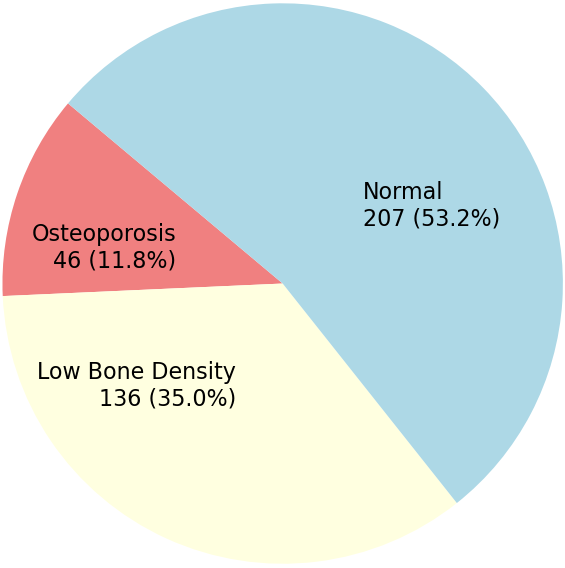}
    \caption{Long-tailed distribution of T-score classifications within the AustinSpine dataset.}
    \label{fig:dist}
\end{figure}

The AustinSpine dataset is a clinically curated collection of spinal CT scans, comprising imaging data from 389 patients, obtained with full ethical approval. Bone density for each scan is quantified using T-scores, a standardized metric widely employed to assess bone health. To ensure the reliability and consistency of annotations, each T-score underwent a thorough review by at least two expert radiologists, significantly enhancing the dataset's inter-rater reliability. Based on the World Health Organization (WHO) criteria for bone mineral density (BMD) \cite{who1994assessment}, the T-scores are categorized into three distinct classes, as detailed in Table \ref{tab:Tscore}. The dataset distribution, visualized in Figure \ref{fig:dist}, reveals a pronounced long-tailed pattern, highlighting the predominance of normal cases relative to the other classifications. This clinically enriched dataset offers a robust and reliable resource for the development and validation of automated bone density prediction models, particularly within real-world clinical settings where precise and consistent annotations are critical.

\begin{table}
\centering
\caption{World Health Organization (WHO) criteria for classification of patients with bone mineral density (BMD) ~\cite{who1994assessment}.}
\resizebox{\columnwidth}{!}{%
\begin{tabular}{c|c|c}
    \hline
    \textbf{T-score Range} & \textbf{Condition} & \textbf{Description} \\
    \hline
    -4 to -2.5 & Osteoporosis & Porous bone that can lead to fractures \\
    -2.5 to -1 & Osteopenia & Low Bone Density \\
    -1 and above & Normal & As compared to an average 30-year-old \\
    \hline
\end{tabular}%
}
\label{tab:Tscore}
\end{table}

\subsection{Evaluation Matrices}

For a fair comparison, we evaluated each method's overall classification performance on the test set using accuracy and ROC AUC scores. Additionally, we assessed sensitivity and specificity to understand how effectively each model handles both minority and majority classes within the long-tailed AustinSpine dataset.

\section{Experiment}

Our experiment is based on CT segmentation technology, utilizing two mainstream segmentation algorithms: CTSpine1K~\cite{deng2021ctspine1k} and TotalSegmentator \cite{wasserthal2023totalsegmentator}. CTSpine1K is a large-scale spinal CT dataset containing 1005 scans with over 11,100 labeled vertebrae, designed to advance research on spine-related image analysis tasks. TotalSegmentator is a deep learning segmentation model capable of automatically segmenting 104 major anatomical structures in CT images, including organs\cite{zhang2024segreg}, bones, muscles, and vessels, with robustness and high accuracy.

We use these algorithms to segment the lumbar vertebra L1 as input. The L1 vertebra, located at the top of the lumbar spine, serves as a critical load-bearing structure, supporting the upper body's weight while allowing flexibility and movement. Its position between the thoracic spine and the lower lumbar vertebrae makes it vital for both structural stability and mobility. Furthermore, the bone mineral density (BMD) of the L1 vertebra plays a crucial role in assessing overall bone health, serving as a key indicator in the diagnosis of osteoporosis and the evaluation of fracture risk ~\cite{ramschutz2024cervicothoracic}.

Through comparative experiments, we found that the segmentation results based on TotalSegmentator consistently outperformed those achieved by CTSpine1K in overall performance. Therefore, we selected the segmentation outputs of TotalSegmentator as the input for MedConv.

\begin{figure}[h]
    \centering
    \includegraphics[width=0.8\linewidth]{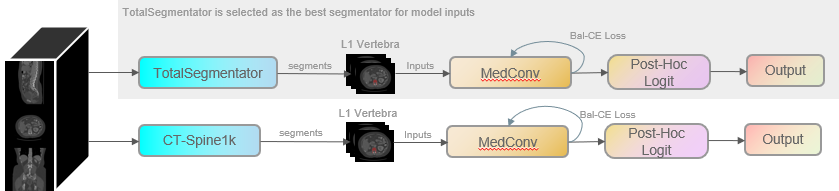}
    \caption{
        Experiment pipeline for evaluating segmentation methods and their impact on downstream tasks. 
        This flowchart illustrates the comparison between CTSpine1K~\cite{deng2021ctspine1k} and TotalSegmentator~\cite{wasserthal2023totalsegmentator}, two widely used segmentation algorithms. 
        Both methods segment the L1 vertebra from input CT images, with the outputs subsequently processed by the MedConv module, 
        followed by post-hoc logits optimized with balanced cross-entropy loss. 
        TotalSegmentator was identified as the superior model, producing more robust and accurate segmentation results, which were selected as inputs for the MedConv module.}
    \label{fig:experiment_flowchart}
\end{figure}

\subsection{Comparative Study}

\begin{table}[htbp]
\vspace{-0.5cm}
    \centering
    \caption{Comparative performance of various models on the given metrics.}
    \resizebox{\linewidth}{!}{%
    \begin{tabular}{l|c|c|c|c|c}
        \hline
        \textbf{Model} & \textbf{Accuracy} & \textbf{Sensitivity} & \textbf{Specificity} & \textbf{F1 Score} & \textbf{ROC AUC} \\
        \hline
        resnet10t.c3\_in1k+pretrain & 58.97 & 58.97 & 79.49 & 59.46 & 67.90 \\
        resnet14t.c3\_in1k+pretrain & 56.41 & 56.41 & 78.21 & 56.53 & 70.12 \\
        resnet18.a1\_in1k & 48.72 & 48.72 & 74.36 & 44.87 & 65.06 \\
        resnet18.a1\_in1k+windows & 47.44 & 47.44 & 73.72 & 44.81 & 63.63 \\
        resnet18.a1\_in1k+balaug & 47.44 & 47.44 & 73.72 & 42.75 & 64.67 \\
        resnet18.a1\_in1k+pretrain & 62.82 & 62.82 & 81.41 & 62.34 & 74.51 \\
        resnet18.a1\_in1k+pretrain+balce & 62.82 & 62.82 & 81.41 & 62.28 & 75.79 \\
        resnet18.a1\_in1k+pretrain+balaug & 57.69 & 57.69 & 78.85 & 55.47 & 75.02 \\
        resnet18.a1\_in1k+pretrain+windows & 56.41 & 56.41 & 78.21 & 54.92 & 69.53 \\
        resnet18.a1\_in1k+pretrain+balaug+windows & 58.97 & 58.97 & 79.49 & 58.64 & 77.71 \\
        resnet34.a1\_in1k & 48.72 & 48.72 & 74.36 & 46.77 & 64.47 \\
        resnet34.a1\_in1k+balaug & 46.15 & 46.15 & 73.08 & 46.41 & 66.15 \\
        resnet34.a1\_in1k+windows & 48.72 & 48.72 & 74.36 & 47.33 & 63.93 \\
        resnet34.a1\_in1k+pretrain & 58.97 & 58.97 & 79.49 & 57.45 & 83.01 \\
        resnet34.a1\_in1k+pretrain+balce & 57.69 & 57.69 & 78.85 & 56.54 & 74.51 \\
        resnet34.a1\_in1k+pretrain+balaug & 56.41 & 56.41 & 78.21 & 56.34 & 73.69 \\
        resnet50.a1\_in1k & 44.87 & 44.87 & 72.44 & 41.66 & 59.94 \\
        resnet50.a1\_in1k+pretrain & 57.69 & 57.69 & 78.85 & 55.89 & 76.87 \\
        resnet50.a1\_in1k+pretrain+balaug & 55.13 & 55.13 & 77.56 & 53.52 & 70.81 \\
        resnet50.a1\_in1k+pretrain+balce & 64.10 & 64.10 & 82.05 & 65.14 & 78.43 \\
        resnet50.a1\_in1k+pretrain+balce+schdulefree & 57.69 & 57.69 & 78.85 & 57.40 & 71.40 \\
        resnet50.a1\_in1k+pretrain+balce+balaug & 62.82 & 62.82 & 81.41 & 61.13 & 76.53 \\
        resnet50.a1\_in1k+pretrain+balce+resample & 61.54 & 61.54 & 80.77 & 60.61 & 73.25 \\
        resnet50.a1\_in1k+pretrain+sam & 55.13 & 55.13 & 77.56 & 54.36 & 73.10 \\
        resnet50.a1\_in1k+pretrain+balce+sam & 56.41 & 56.41 & 78.21 & 55.82 & 73.22 \\
        resnet50.a1\_in1k+trainParams+balaug & 53.85 & 53.85 & 76.92 & 53.13 & 74.58 \\
        mobilenetv2\_100.ra\_in1k+pretrain & 52.56 & 52.56 & 76.28 & 51.89 & 69.26 \\
        mobilenetv2\_100.ra\_in1k+pretrain+balce & 56.41 & 56.41 & 78.21 & 55.45 & 71.52 \\
        efficientnet\_b0.ra\_in1k+pretrain & 57.69 & 57.69 & 78.85 & 56.02 & 74.14 \\
        efficientnet\_b0.ra\_in1k+pretrain+balce & 60.26 & 60.26 & 80.13 & 58.73 & 78.06 \\
        resnext50\_32x4d.a1h\_in1k+pretrain & 60.26 & 60.26 & 80.13 & 60.77 & 76.06 \\
        resnext50\_32x4d.a1h\_in1k+pretrain+balce & 55.13 & 55.13 & 77.56 & 52.64 & 71.28 \\
        resnext50\_32x4d.a1h\_in1k+pretrain+balaug & 50.00 & 50.00 & 75.00 & 46.06 & 62.15 \\
        resnet101.a1\_in1k+pretrain & 55.13 & 55.13 & 77.56 & 54.78 & 73.30 \\
        resnet101.a1\_in1k+pretrain+balce & 58.97 & 58.97 & 79.49 & 55.27 & 73.30 \\
        resnet101.a1\_in1k+pretrain+balce+sam & 55.13 & 55.13 & 77.56 & 47.70 & 71.06 \\
        resnet152.tv\_in1k+pretrain & 50.00 & 50.00 & 75.00 & 47.91 & 69.06 \\
        resnet152.tv\_in1k+pretrain+balce & 57.69 & 57.69 & 78.85 & 58.30 & 70.76 \\
        resnet152.tv\_in1k+pretrain+balce+sam & 52.56 & 52.56 & 76.28 & 48.72 & 69.82 \\ 
        ViT+pretrain & 33.54 & 33.54 & 66.77 & 17.93 & 58.35 \\ 
        JointViT + pretrain & 41.03 & 41.03 & 76.92 & 53.85 & 60.78 \\
        JointViT +pretrain + balce & 43.59 & 43.59 & 71.79 & 34.60 & 56.81 \\ \hline
        \textbf{MedConv (Ours)} & \textbf{65.38} & \textbf{65.38} & \textbf{82.69} & \textbf{66.37} & \textbf{79.34} \\
        \hline
    \end{tabular}}
    \label{tab:comparative_results}
\end{table}

In this comparative experiment, we evaluated various models based on their performance metrics, including accuracy, sensitivity, specificity, F1 score, and ROC AUC. All models were tested using the segmentation outputs of TotalSegmentator, which were selected due to their superior performance in our preliminary ablation studies.

The results indicate that our proposed MedConv model achieved the highest accuracy of 65.38, surpassing other models such as resnet50.a1 in1k+pretrain+balce, which scored 64.10, and resnet34.a1 in1k+pretrain, which achieved an accuracy of 58.97. This demonstrates the effectiveness of MedConv in handling complex medical imaging data.

In terms of sensitivity and specificity, the MedConv model demonstrated remarkable results, with scores of 65.38 and 82.69, respectively. These metrics highlight MedConv's capability to accurately identify positive cases while minimizing the occurrence of false positives. In comparison, the next highest sensitivity was achieved by resnet50.a1 in1k+pretrain+balce with a score of 64.10, emphasizing MedConv's superior ability to distinguish true positives and true negatives with greater precision.

The F1 score for MedConv is 66.37, further establishing its robustness in balancing precision and recall. This metric is particularly critical in medical applications where both false positives and false negatives can significantly affect diagnostic reliability. MedConv's performance in this regard surpasses many other models, reinforcing its suitability for high-stakes scenarios where precise predictions are essential.

The ROC AUC for MedConv is 79.34, reflecting its overall performance across various classification thresholds. This metric is crucial for clinical applications, where decision-making often relies on evaluating a model's behavior across different thresholds. MedConv's high ROC AUC score highlights its reliability and effectiveness in real-world medical applications.

\begin{table}[htbp]
    \centering
    \caption{Comparative performance of CTspine1K and TotalSegmentator with different inputs.}
    \resizebox{\linewidth}{!}{%
    \begin{tabular}{l|l|c|c|c|c|c}
        \hline
        \textbf{Model} & \textbf{Input} & \textbf{Accuracy} & \textbf{Sensitivity} & \textbf{Specificity} & \textbf{F1 Score} & \textbf{ROC AUC} \\
        \hline
        \multirow{2}{*}{mobilenetv2\_100.ra\_in1k} 
            & CTspine1K        & 39.74 & 39.74 & 69.87 & 47.71 & 36.03 \\ 
            & TotalSegmentator & 52.56 & 52.56 & 76.28 & 51.89 & 69.26 \\ 
        \hline
        \multirow{2}{*}{resnet18.a1\_in1k} 
            & CTspine1K        & 46.15 & 46.15 & 73.08 & 41.30 & 53.43 \\ 
            & TotalSegmentator & 48.72 & 48.72 & 74.36 & 44.87 & 65.06 \\ 
        \hline
        \multirow{2}{*}{resnet34.a1\_in1k} 
            & CTspine1K        & 42.31 & 42.31 & 71.15 & 36.01 & 53.60 \\ 
            & TotalSegmentator & 48.72 & 48.72 & 74.36 & 46.77 & 64.47 \\ 
        \hline
        \multirow{2}{*}{resnet50.a1\_in1k} 
            & CTspine1K        & 44.87 & 44.87 & 72.44 & 37.97 & 59.32 \\ 
            & TotalSegmentator & 44.87 & 44.87 & 72.44 & 41.66 & 59.94 \\ 
        \hline
        \multirow{2}{*}{resnet101.a1\_in1k} 
            & CTspine1K        & 43.59 & 43.59 & 71.79 & 34.97 & 57.91 \\ 
            & TotalSegmentator & 55.13 & 55.13 & 77.56 & 54.78 & 73.30 \\ 
        \hline
        \multirow{2}{*}{resnet152.tv\_in1k} 
            & CTspine1K        & 42.31 & 42.31 & 71.15 & 36.10 & 54.29 \\ 
            & TotalSegmentator & 50.00 & 50.00 & 75.00 & 47.91 & 69.06 \\ 
        \hline
    \end{tabular}}
    \label{tab:comparative_results0}
\end{table}

To ensure the robustness of our experimental results, we performed additional evaluations using the same models but with segmentation outputs generated by CTspine1K instead of TotalSegmentator. CTspine1K, a specialized segmentation tool for spine imaging, serves as an alternative segmentation source. However, as shown in Table~\ref{tab:comparative_results0}, models consistently underperformed when using CTspine1K outputs compared to those using TotalSegmentator. Key metrics, including accuracy, sensitivity, specificity, F1 score, and ROC AUC, exhibited significant declines across all models.
Notably, the results demonstrate that TotalSegmentator’s high-quality and comprehensive segmentation is pivotal for achieving superior model performance. Furthermore, when TotalSegmentator outputs were used, model performance either improved or remained stable as model parameter counts increased. This trend highlights the richness of the information provided by TotalSegmentator, which facilitates more effective utilization of complex model architectures.
Based on these findings, we selected TotalSegmentator as the default segmentation input source for all subsequent experiments to ensure consistency and optimize the models' potential.

In summary, these results demonstrate that the MedConv model not only outperforms other tested alternatives but also represents a significant advancement in model architecture for medical imaging tasks. By leveraging the high-quality segmentation results from TotalSegmentator, MedConv has shown exceptional accuracy, sensitivity, specificity, and overall robustness. These findings underscore the potential of MedConv to enhance diagnostic accuracy and improve patient outcomes, making it a promising tool for clinical and medical research applications.

\subsection{Ablation Study}

This section presents a comprehensive analysis of the proposed approach through three separate experiments. The first experiment focuses on validating the effectiveness of BalCE loss across different backbone architectures, highlighting its role in addressing class imbalance and improving overall performance. The second experiment investigates the influence of the hyperparameter \(\tau_1\), which balances the loss contribution from positive and negative samples, on key performance metrics. This analysis aims to identify the optimal value of \(\tau_1\) for achieving stable and robust performance. Finally, the third experiment evaluates the sensitivity of the model to variations in the hyperparameter \(\tau_2\), which serves as an ad-hoc weighting parameter within the MedConv framework. These experiments collectively underscore the robustness, adaptability, and fine-tuning flexibility of the proposed method in addressing challenges associated with class imbalance in medical imaging.

\subsubsection{Impact of BalCE Loss on Different Backbones}

\begin{table}[h!]
\centering
\caption{Ablation Study: Comparison of BalCE Loss Across Different Backbones}
\resizebox{\columnwidth}{!}{
\begin{tabular}{l|c|c|c|c|c}
\hline
\textbf{Model}          & \textbf{Accuracy}       & \textbf{Sensitivity}    & \textbf{Specificity}    & \textbf{F1 Score}       & \textbf{ROC AUC}        \\ \hline
mobilenetv2 w/o         & 52.56                  & 52.56                  & 76.28                  & 51.89                  & 69.26                  \\
mobilenetv2 w/          & 56.41 \textcolor{green}{(+3.85)} & 56.41 \textcolor{green}{(+3.85)} & 78.21 \textcolor{green}{(+1.93)} & 55.45 \textcolor{green}{(+3.56)} & 71.52 \textcolor{green}{(+2.26)} \\\hline
efficientnet w/o        & 57.69                  & 57.69                  & 78.85                  & 56.02                  & 74.14                  \\
efficientnet w/         & 60.26 \textcolor{green}{(+2.57)} & 60.26 \textcolor{green}{(+2.57)} & 80.13 \textcolor{green}{(+1.28)} & 58.73 \textcolor{green}{(+2.71)} & 74.14                  \\\hline
resnet34 w/o            & 58.97                  & 58.97                  & 79.49                  & 57.45                  & 83.01                  \\
resnet34 w/             & 57.69 \textcolor{red}{(-1.28)} & 57.69 \textcolor{red}{(-1.28)} & 78.85 \textcolor{red}{(-0.64)} & 56.54 \textcolor{red}{(-0.91)} & 74.51 \textcolor{red}{(-8.50)} \\\hline
resnet50 w/o            & 57.69                  & 57.69                  & 78.85                  & 55.89                  & 76.87                  \\
\textbf{resnet50 w/}             & \textbf{64.10} \textcolor{green}{(+6.41)} & \textbf{64.10} \textcolor{green}{(+6.41)} & \textbf{82.05} \textcolor{green}{(+3.20)} & \textbf{65.14} \textcolor{green}{(+9.25)} & \textbf{78.43} \textcolor{green}{(+1.56)} \\\hline
resnet101 w/o           & 55.13                  & 55.13                  & 77.56                  & 54.78                  & 73.30                  \\
resnet101 w/            & 58.97 \textcolor{green}{(+3.84)} & 58.97 \textcolor{green}{(+3.84)} & 79.49 \textcolor{green}{(+1.93)} & 55.27 \textcolor{green}{(+0.49)} & 73.30 \textcolor{red}{(+0.00)} \\\hline
resnet152 w/o           & 50.00                  & 50.00                  & 75.00                  & 47.91                  & 69.06                  \\
resnet152 w/            & 57.69 \textcolor{green}{(+7.69)} & 57.69 \textcolor{green}{(+7.69)} & 78.85 \textcolor{green}{(+3.85)} & 58.30 \textcolor{green}{(+10.39)} & 70.76 \textcolor{green}{(+1.70)} \\\hline
\end{tabular}
}
\label{tab:ablation}
\end{table}

The results of integrating BalCE loss across different backbones are summarized in Table \ref{tab:ablation}. Consistent performance improvements are observed across most architectures, with significant gains in accuracy, sensitivity, and F1 score. Notably, ResNet50 exhibits the highest improvement, achieving a 6.41\% increase in accuracy and a 9.25\% increase in F1 score. These findings underscore the effectiveness of BalCE loss in addressing data imbalance, particularly in challenging medical imaging scenarios. However, a minor performance drop is noted in ResNet34, potentially due to overfitting or incompatibility between the backbone and loss function.

\subsubsection{Effect of \(\tau_1\) on Model Performance}

\begin{table}[h]
\centering
\caption{Ablation study of different \(\tau_1\) hyperparameter settings and their impact on model performance metrics.}
\resizebox{\columnwidth}{!}{
\begin{tabular}{c|c|c|c|c|c}
\hline
\textbf{$\tau_1$} & \textbf{Accuracy} & \textbf{Sensitivity} & \textbf{Specificity} & \textbf{F1} & \textbf{AUC} \\\hline
0     & 57.69   & 57.69   & 78.85   & 55.89   & 76.87   \\
0.25  & 58.97 {\color{green}(+1.28)} & 58.97 {\color{green}(+1.28)} & 79.49 {\color{green}(+0.64)} & 59.25 {\color{green}(+3.36)} & 75.42 {\color{red}(-1.45)} \\
0.5   & 57.69 {\color{green}(+0.00)} & 57.69 {\color{green}(+0.00)} & 78.85 {\color{green}(+0.00)} & 57.77 {\color{green}(+1.88)} & 78.33 {\color{green}(+1.46)} \\
0.65  & 58.97 {\color{green}(+1.28)} & 58.97 {\color{green}(+1.28)} & 79.49 {\color{green}(+0.64)} & 58.07 {\color{green}(+2.18)} & 70.69 {\color{red}(-6.18)} \\
0.75  & 61.54 {\color{green}(+3.85)} & 61.54 {\color{green}(+3.85)} & 80.77 {\color{green}(+1.92)} & 61.61 {\color{green}(+5.72)} & 77.12 {\color{green}(+0.25)} \\
0.85  & 61.54 {\color{green}(+3.85)} & 61.54 {\color{green}(+3.85)} & 80.77 {\color{green}(+1.92)} & 60.12 {\color{green}(+4.23)} & 74.11 {\color{red}(-2.76)} \\
0.9   & 58.97 {\color{green}(+1.28)} & 58.97 {\color{green}(+1.28)} & 79.49 {\color{green}(+0.64)} & 57.83 {\color{green}(+1.94)} & 74.38 {\color{red}(-2.49)} \\
0.92  & 52.56 {\color{red}(-5.13)} & 52.56 {\color{red}(-5.13)} & 76.28 {\color{red}(-2.57)} & 50.29 {\color{red}(-5.60)} & 70.88 {\color{red}(-5.99)} \\
0.95  & 61.54 {\color{green}(+3.85)} & 61.54 {\color{green}(+3.85)} & 80.77 {\color{green}(+1.92)} & 60.66 {\color{green}(+4.77)} & 74.46 {\color{red}(-2.41)} \\
0.96  & 56.41 {\color{red}(-1.28)} & 56.41 {\color{red}(-1.28)} & 78.21 {\color{red}(-0.64)} & 57.12 {\color{green}(+1.23)} & 73.40 {\color{red}(-3.47)} \\
0.97  & 57.69 {\color{green}(+0.00)} & 57.69 {\color{green}(+0.00)} & 78.85 {\color{green}(+0.00)} & 56.52 {\color{green}(+0.63)} & 69.30 {\color{red}(-7.57)} \\
0.99  & 53.85 {\color{red}(-3.84)} & 53.85 {\color{red}(-3.84)} & 76.92 {\color{red}(-1.93)} & 50.72 {\color{red}(-5.17)} & 70.09 {\color{red}(-6.78)} \\
0.995 & 58.97 {\color{green}(+1.28)} & 58.97 {\color{green}(+1.28)} & 79.49 {\color{green}(+0.64)} & 59.49 {\color{green}(+3.60)} & 72.61 {\color{red}(-4.26)} \\
0.999 & 60.26 {\color{green}(+2.57)} & 60.26 {\color{green}(+2.57)} & 80.13 {\color{green}(+1.28)} & 59.36 {\color{green}(+3.47)} & 78.43 {\color{green}(+1.56)} \\
\textbf{1}     & \textbf{64.10} {\color{green}(+6.41)} & \textbf{64.10} {\color{green}(+6.41)} & \textbf{82.05} {\color{green}(+3.20)} & \textbf{65.14} {\color{green}(+9.25)} & \textbf{78.43} {\color{green}(+1.56)} \\
1.1   & 57.69 {\color{green}(+0.00)} & 57.69 {\color{green}(+0.00)} & 78.85 {\color{green}(+0.00)} & 57.88 {\color{green}(+1.99)} & 74.98 {\color{red}(-1.89)} \\
1.5   & 56.41 {\color{red}(-1.28)} & 56.41 {\color{red}(-1.28)} & 78.21 {\color{red}(-0.64)} & 56.45 {\color{red}(-0.56)} & 75.76 {\color{red}(-1.11)} \\
2     & 52.56 {\color{red}(-5.13)} & 52.56 {\color{red}(-5.13)} & 76.28 {\color{red}(-2.57)} & 45.26 {\color{red}(-10.63)} & 74.73 {\color{red}(-2.14)} \\\hline
\end{tabular}
}
\label{tab:ablation_tau1}
\end{table}

The ablation study under the default condition of \(\tau_1 = \tau_2\) demonstrates that the model achieves its best performance when \(\tau_1 = 1\). This setting effectively balances the loss function, addressing the challenges posed by class imbalance and enhancing model robustness. The analysis confirms that \(\tau_1 = 1\) is the optimal choice for achieving a stable trade-off across performance metrics, providing a strong baseline for further exploration. Subsequently, additional ablations focus on varying \(\tau_2\) while keeping \(\tau_1\) fixed at its optimal value, allowing for a more comprehensive evaluation of the proposed approach.

\subsubsection{Effect of \(\tau_2\) on Model Performance}

We further evaluate the impact of varying the hyperparameter \(\tau_2\) on model performance. This experiment leverages segmentations generated by TotalSegmentator as input, exploring the sensitivity of key performance metrics to changes in \(\tau_2\).

\begin{table}[h]
\centering
\caption{Ablation study of different \(\tau_2\) hyperparameter settings and their impact on model performance metrics.}
\begin{tabular}{c|c|c|c|c|c}
\hline
\textbf{$\tau_2$} & \textbf{Accuracy} & \textbf{Sensitivity} & \textbf{Specificity} & \textbf{F1} & \textbf{AUC} \\\hline
1.0   & 0.6410 & 0.6410 & 0.8205 & 0.6514 & 0.7843 \\
0.9   & 0.6410 & 0.6410 & 0.8205 & 0.6514 & 0.7870 \\
0.8   & 0.6410 & 0.6410 & 0.8205 & 0.6514 & 0.7877 \\
0.7   & 0.6410 & 0.6410 & 0.8205 & 0.6514 & 0.7894 \\
0.6   & 0.6538 & 0.6538 & 0.8269 & 0.6637 & 0.7919 \\
\textbf{0.5}   & \textbf{0.6538} & \textbf{0.6538} & \textbf{0.8269} & \textbf{0.6637} & \textbf{0.7934} \\
0.4   & 0.6410 & 0.6410 & 0.8205 & 0.6493 & 0.7951 \\
0.3   & 0.6410 & 0.6410 & 0.8205 & 0.6493 & 0.7961 \\
0.2   & 0.6282 & 0.6282 & 0.8141 & 0.6359 & 0.7971 \\
0.1   & 0.6282 & 0.6282 & 0.8141 & 0.6337 & 0.7986 \\\hline
\end{tabular}
\label{tab:ablation_tau2}
\end{table}

\begin{figure}[h]
\centering
\includegraphics[width=\columnwidth]{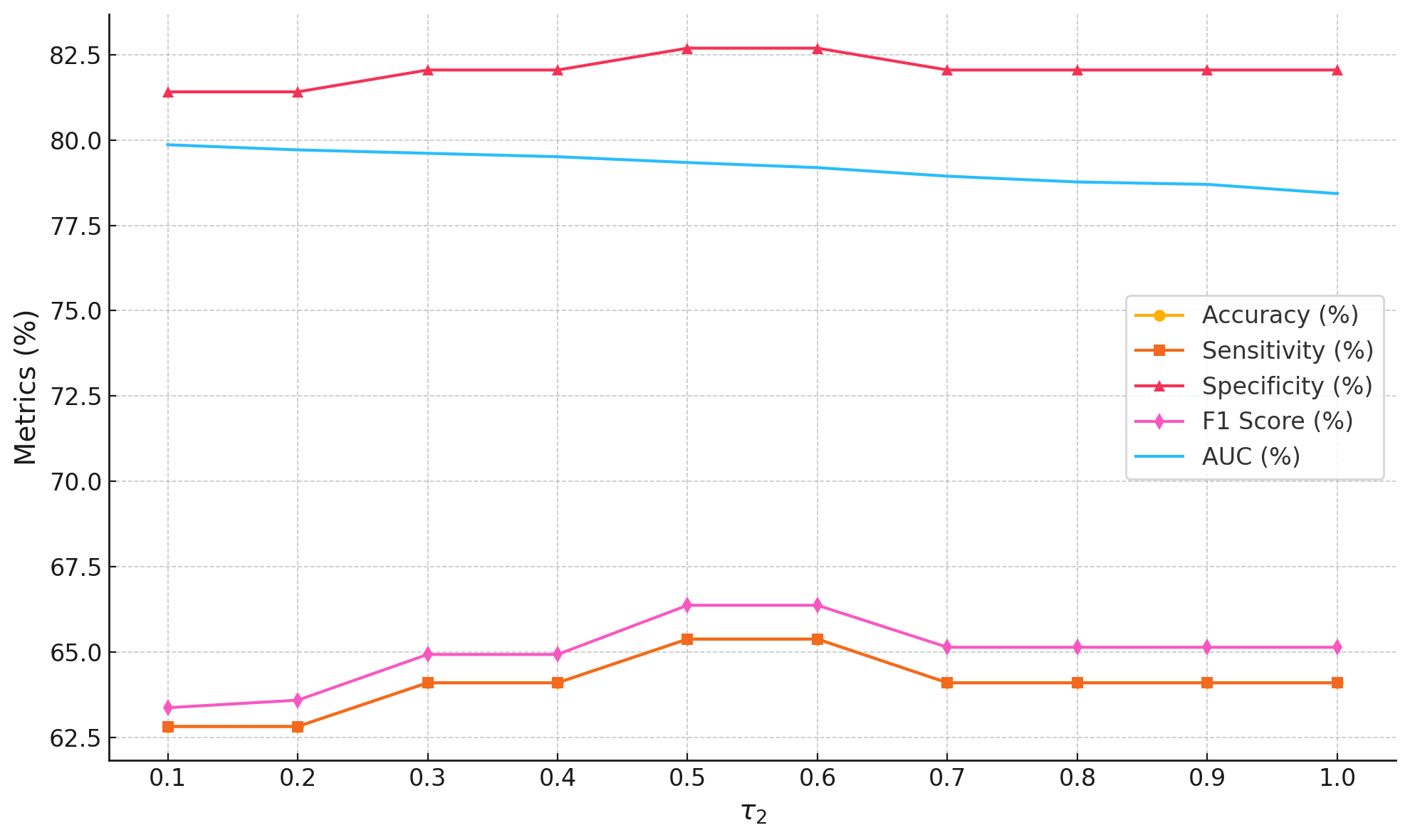}
\caption{Ablation study showing the impact of different $\tau_2$ settings on model performance metrics. Each line represents a distinct metric: Accuracy, Sensitivity, Specificity, F1 Score, and AUC.}
\label{fig:ablation_tau2}
\end{figure}

As shown in Table \ref{tab:ablation_tau2} and Figure \ref{fig:ablation_tau2}, the model maintains stable performance for \(\tau_2\) values between 1.0 and 0.7, with accuracy, sensitivity, and specificity hovering around 64.10\%. A significant improvement is observed at \(\tau_2 = 0.6\) and \(\tau_2 = 0.5\), where accuracy rises to 65.38\% and F1 score reaches 66.37\%. This suggests that moderate \(\tau_2\) values balance precision and recall effectively.

When \(\tau_2\) is reduced further, a decline in performance becomes evident. At \(\tau_2 = 0.1\), accuracy drops to 62.82\%, with corresponding decreases in sensitivity and specificity. These findings highlight the importance of tuning \(\tau_2\) to achieve optimal results, emphasizing its role in improving model robustness and generalization.

\section{Conclusion}

In this study, we introduced MedConv, a convolutional neural network designed for bone density prediction via CT scans. MedConv outperforms transformer-based methods in accuracy, sensitivity, and specificity, while maintaining a significantly lower computational cost. By employing a 3D ResNet-50 backbone, the model effectively captures volumetric spatial information, which is critical for precise bone health assessment. This capability enables MedConv to be more suited for practical applications in clinical and resource-constrained settings compared to transformer models.

To address the inherent challenges of imbalanced and long-tailed datasets in real-world medical imaging, we adopted a Balanced Cross-Entropy (Bal-CE) loss function combined with post-hoc logit adjustment techniques. These strategies demonstrated robust improvements in classification accuracy and model calibration, as evidenced by the performance gains observed in our experiments on the AustinSpine dataset. Specifically, MedConv achieved a 21\% improvement in classification accuracy and a 20\% increase in ROC AUC compared to prior state-of-the-art methods, solidifying its position as a benchmark in this domain.

The ablation studies further emphasized the importance of hyperparameter tuning in optimizing model performance. For the logit adjustment hyperparameter $\tau_1$, the results indicate that the optimal setting of $\tau_1 = 1$ provides a balanced trade-off across various performance metrics, achieving the highest accuracy and F1 score. Similarly, an ad-hoc analysis of $\tau_2$ revealed that moderate values, particularly $\tau_2 = 0.5$, yielded significant performance gains. The model exhibited improved robustness and generalization at $\tau_2 = 0.5$, with accuracy increasing to 65.38\% and F1 score reaching 66.37\%. This suggests that $\tau_2$ plays a crucial role in calibrating the relative contributions of minority and majority classes, thereby enhancing overall performance.

Additionally, the study underscores the importance of high-quality segmentation tools such as TotalSegmentator, which played a pivotal role in enhancing the overall performance of MedConv. The segmentation outputs from TotalSegmentator provided superior input quality, enabling MedConv to better leverage the volumetric spatial information for accurate predictions.

MedConv’s success in balancing computational efficiency and predictive performance highlights its potential for broader applications in clinical settings, where timely and accurate diagnoses are imperative. Future work may explore extending MedConv to other imaging modalities and clinical tasks, as well as further refining its architecture to enhance versatility and scalability in diverse healthcare environments. By bridging the gap between advanced deep learning techniques and practical deployment, MedConv sets a promising foundation for improved diagnostic tools in the fight against osteoporosis and other bone health conditions.

\end{document}